\def\ArtDir{main/}
\documentclass{ws-ijprai}
\def\ArtDir{main/}

\usepackage{xcolor}
\usepackage{tabularx,booktabs}
\usepackage{makecell}
\newcolumntype{C}{>{\centering\arraybackslash}X}
\usepackage{multirow}
\usepackage{caption}
\usepackage{subcaption}
\usepackage{rotating}

\usepackage{lscape}

\begin{document}

\markboth{De Bois, El Yacoubi, Ammi}{Enhance the Interpretability of Deep Models in Heathcare Through Attention}

%
\catchline{}{}{}{}{}
%
\title{Enhancing the Interpretability of Deep Models in Heathcare Through Attention: Application to Glucose Forecasting for Diabetic People}

\author{Maxime De Bois}

\address{CNRS-LIMSI and Universit\'{e} Paris-Saclay\\
Orsay, France\\
\email{maxime.debois@limsi.fr}}

\author{Moun{\^i}m A. El Yacoubi}

\address{Samovar, CNRS, T{\'e}l{\'e}com SudParis, Institut Polytechnique de Paris\\
\'{E}vry, France\\}

\author{Mehdi Ammi}

\address{Universit\'{e} Paris 8\\
Saint-Denis, France\\}


\maketitle


\begin{abstract}

The adoption of  deep learning in healthcare is hindered by their ``black box'' nature. In this paper, we explore the RETAIN architecture for the task of glusose forecasting for diabetic people. By using a two-level attention mechanism, the recurrent-neural-network-based RETAIN model is interpretable. We evaluate the RETAIN model on the type-2 IDIAB and the type-1 OhioT1DM datasets by comparing its statistical and clinical performances against two deep models and three models based on decision trees. We show that the RETAIN model offers a very good compromise between accuracy and interpretability, being almost as accurate as the LSTM and FCN models while remaining interpretable. We show the usefulness of its interpretable nature by analyzing the contribution of each variable to the final prediction. It revealed that signal values older than one hour are not used by the RETAIN model for the 30-minutes ahead of time prediction of glucose. Also, we show how the RETAIN model changes its behavior upon the arrival of an event such as carbohydrate intakes or insulin infusions. In particular, it showed that the patient's state before the event is particularily important for the prediction. Overall the RETAIN model, thanks to its interpretability, seems to be a very promissing model for regression or classification tasks in healthcare.

\end{abstract}

\section{Introduction}

One of the major obstacles to the adoption of deep learning in the medical field is the lack of interpretability of models, often characterized as ``black boxes'' \cite {ching2018opportunities}. This need for interpretability can be explained by several distinct reasons \cite {adadi2018peeking}. The main reason is undoubtedly the need to understand the decisions made by the models in order to be able to trust them. This need is increased when the predictions are unexpected and may threaten the life of the patient. Indeed, although statistically verified, a decision may be the result of a bias of the model in the training data. For example, by training a model predicting the probability of death of patients from pneumonia, Ba \textit {et al.} showed that their model associated asthma patients with a low probability of death \cite {ba2014deep}. This erroneous association made by the model came from a bias in the training data. Indeed, asthma patients are often treated first in hospitals, resulting in low statistical mortality. The second reason behind the need for interpretable models is that they can improve our general knowledge of pathologies.

In general, interpretable models, such as linear regression or decision trees, perform poorly in comparison with more complex models. This increased complexity, associated with a gain in performance, very often leads to a large drop in interpretability (e.g., deep neural networks, random forests). Thus, recent endeavors have been focused on how to interpret complex models, and in particular, deep models. Among these efforts, we can identify two different approaches. The first one aims to measure, visualize, the importance of the input data on the predictions. For example, Simonyan \textit {et al.} proposed the construction of a saliency map that identifies the important pixels for the classification of images \cite {simonyan2013deep}. These saliency maps were used by Ma \textit {et al.} to analyze the nature of adversarial attacks on a convolutional network trained on medical images \cite {ma2020understanding}. In their work, Lundberg \textit {et al.} proposed a framework aiming to measure the importance of each input feature on the predictions \cite {lundberg2017unified}, with its application for the prevention of hypoxemia during surgeries \cite {lundberg2018explainable}. Rather than proposing methods for interpreting black boxes, a large number of researchers are looking at architectural modifications making deep models directly more interpretable. Among these new architectures, the most notable are based on the innovative mechanism of attention. It was introduced by Bahdanau \textit {et al.} in the field of machine translation \cite {bahdanau2014neural} and then used in the Transformer architecture \cite {vaswani2017attention}. Thanks to the attention mechanism, models based on the Transformer architecture are today the models obtaining the best results for all tasks relating to automatic language processing. Built for models using sequential data, the principle of attention allows the model to focus on one or more parts of the sequence in order to make its prediction. Allowing to obtain equal or even better performance in certain fields, the attention paid by the model to the different temporal instants is quantifiable, improving the interpretability of the model. The general attention mechanism has many variants, such as the Transformer's multi-head attention or the two-level attention of the RETAIN model. The latter was proposed by Choi \textit {et al.} in order to process and analyze electronic health records \cite {choi2016retain}. Its temporal attention coupled with its attention to the variable allows it to directly quantify the contribution of each variable, at each moment, to the final prediction.

In this paper we explore the use of the RETAIN architecture and the attention mechanism for the forecasting of future glucose values for diabetic people. These people have difficulties regulating their blood glucose level because of the non-production of insulin by their pancreas (type-1 diabetes) or because of the increasing body resistance to its action (type-2 diabetes). Predicting future glucose values can help them avoid short-term (e.g., coma) and long-term (e.g., cardiovascular diseases) complications induced by hypoglycemia (glucose level below 70 mg/dL) and hyperglycemia (glucose level above 180 mg/dL). Thanks to the increasing availability of diabetes-related data, the field of glucose prediction has moves away from traditionnal simple autoregressive models \cite{sparacino2007glucose} to more complex machine learning or deep learning models. In particular, recurrent neural networks have recently generated a lot of interest because of their temporal nature, making them particularily suitable for the task of predicting future glucose values \cite{mirshekarian2017using,sun2018predicting,martinsson2019blood,debois2019prediction}. As time-series can be seen as one-dimension images, convolutional neural networks, which are very popular in the image recognition community, have also been tried out for the forecasting of future glucose values with very promissing results \cite{zhu2018deep,li2019convolutional,de2020adversarial}. While these models are more accurate, they are also less interpretable. Interpretability is very important in healthcare fields and especially in glucose prediction. Indeed, it allows the patient to make more informed decisions based on the model's predictions (e.g., why is the model predicting a future hypoglycemia?). Also this can improve the understanding of the individual specificities of the disease by either him/her or the doctor. Finally, it can help the scientist in the construction of the model, whether in its architecture or in the nature of the data used. While not fully interpretable, some researchers explored the use of decision-tree based models such as random forests or gradient boosting machines \cite{midroni2018predicting,jeon2019predicting,mayo2019glycemic}.

Our contributions are the following:\begin{itemize}
    \item The RETAIN architecture has initially been proposed in the context of heart failure detection from electronic health records. As this is a classification task, we adapted the framework for regression problems. We further improve the interpretability of the model by proposing a new metric, the absolute normalized contribution of input variables on the prediction. Compared to the standard contribution metric, it enables the use of statistical tools for the analysis of the model's behavior.
    \item We demonstrate empirically the interest of the RETAIN model by applying it to the challenging task of glucose forecasting for diabetic people. For this purpose we used two different datasets, namely the IDIAB dataset and the OhioT1DM dataset. While the OhioT1DM dataset has been released by Marling \textit{et al.} \cite{marling2018ohiot1dm} and is made of 6 type-1 diabetic people, the IDIAB dataset has been collected by ourseleves and comprises 6 type-2 diabetic patients. We compared the statistical and clinical performances of the RETAIN model against several reference models including deep models and decision-tree-based models. Finally, we demonstrate the usefullness of the interpretability of the RETAIN model by analyzing its behavior predicting future glucose values.
    \item Compared to our previous publication on the topic \cite{debois2020interpreting}, we strengthened the evaluation by adding another dataset, by adding several other reference models, and by personalizing all the models to the diabetic patient.
    \item We open-sourced the source code of this study in a GitHub repository \cite{debois2018retain}. It includes the whole data pipeline and the implementation of the models.
\end{itemize}

This paper is organized as follows. First, we present the attention mechanism, the RETAIN architecture as well as the process of interpreting its predictions. After having detailed the general methodology that was followed in this study, we present the experimental results of the models. Finally, we empirically demonstrate the interest of the RETAIN architecture through various visualization tools for blood glucose prediction.
\section{The Attention Mechanism and the RETAIN Architecture}

This section introduces the attention mechanism and the RETAIN architecture with its two-level attention and the computation of the input variables contribution to the predictions.

\subsection{Presentation of the Attention Mechanism for Regression Problems}

Before describing the RETAIN architecture, we propose to lay the foundations of the attention mechanism applied to regression tasks. The attention mechanism was first introduced in the field of machine translation by Bahdanau \textit{et al.} \cite{bahdanau2014neural}. This field is characterized by the use of sequential multi-input multi-output architectures, the input and output data being represented by vectors of words forming sentences (e.g., translation of a sentence in French into a sentence in English). On the other hand, most regression tasks have only one output resulting in a simplification of the architecture implementing the attention mechanism \cite{ran2019lstm}.


\begin{figure*}[!ht]
    \centering
    \includegraphics[width=0.6\textwidth]{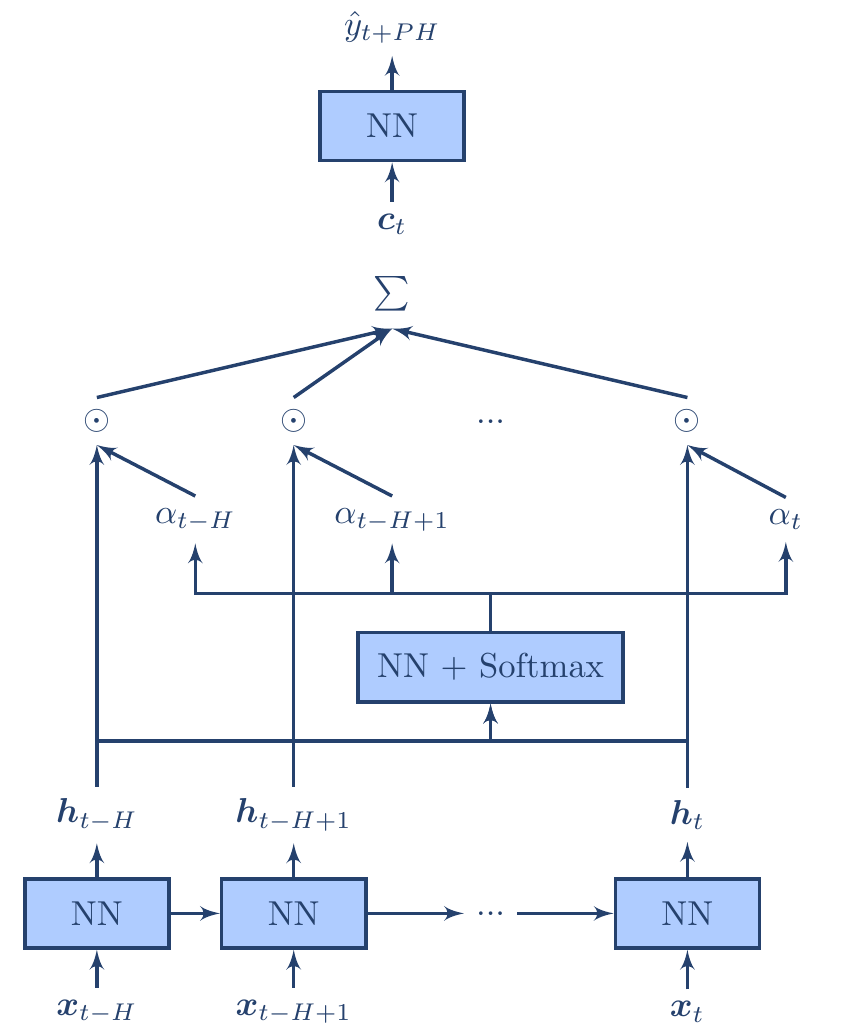}
    \caption{Recurrent neural network using the standard attention mechanism.}
    \label{fig:attention_standard}
\end{figure*}

In the following paragraphs, supported by Figure \ref{fig:attention_standard}, we proceed to describ the architecture of a recurrent neural network implementing the standard attention mechanism. This model aims at predicting the value $ y_ {t + PH} $ from the input data $ \boldsymbol {x} _ {t-H}, ..., \boldsymbol {x} _t $, where $ \boldsymbol { x} _i \in \mathcal {R} ^ r $, $ i \in [t-H, t] $ and $ H $ represents the length of the input sequence.


First, according to Equation \ref{eqn:retain_rrn_standard}, a recurrent neural network RNN transforms the input data into hidden representations $ \boldsymbol {h} _i \in \mathcal {R} ^ p, $ where $ p $ is the number of neurons (or LSTM units for instance) of the recurrent neural network.


\begin{equation}
    \boldsymbol{h}_{t-H}, ..., \boldsymbol{h}_t = \text{RNN}(\boldsymbol{x}_{t-H}, ..., \boldsymbol{x}_t)
    \label{eqn:retain_rrn_standard}
\end{equation}

From the hidden representations $ \boldsymbol {h} _i $, the attention weights $ \alpha_i \in \mathcal {R} $ can be computed according to Equation \ref{eqn:retain_standard_attention}. Equation \ref{eqn:e_i} implements a dense layer of neurons (weights $ \boldsymbol {w} _ \alpha \in \mathcal {R} ^ p $ and bias $ b_ \alpha \in \mathcal { R} $) to compute the relative attention $ e_i \in \mathcal {R} $ of each hidden representation. These attention weights are then normalized into $ \alpha_i $ through the Softmax operation described by Equation \ref {eqn:softmax}. This normalization guarantees attention weights between 0 and 1 and whose sum equals 1.


\begin{subequations}
 \begin{alignat}{2}
e_i & = \boldsymbol{w}_\alpha^T\boldsymbol{h}_i + b\label{eqn:e_i}\\
    \alpha_i & = \frac{\exp{(e_{i})}}{\sum_{j=t-H}^t \exp{(e_j)}} \label{eqn:softmax}
 \end{alignat}\label{eqn:retain_standard_attention}
\end{subequations}

Then, according to Equation \ref{eqn:standard_ctx}, the context vector $ \boldsymbol {c} _t \in \mathcal {R} ^ p $ is computed as the sum of the hidden representations $ \boldsymbol {h} _i $ weighted by their respective attention weight $ \alpha_i $.


\begin{equation}
    \boldsymbol{c}_t = \sum_{i=t-H}^t \alpha_i \boldsymbol{h}_i
    \label{eqn:standard_ctx}
\end{equation}

Finally, according to Equation \ref{eqn:standard_pred}, the model prediction can be computed by a dense layer of neurons (weights $ \boldsymbol {w} _ {out} \in \mathcal {R} ^ p $ and bias $ b_ {out} \in \mathcal {R } $) taking as input the context vector $ \boldsymbol {c} _t $.


\begin{equation}
    \hat{y}_{t+PH} = \boldsymbol{w}_{out}^T \boldsymbol{c}_t + b_{out}
    \label{eqn:standard_pred}
\end{equation}

In comparison with a standard recurrent-neural-network-based architecture, this architecture weights the hidden representations $ \boldsymbol {h} _i $ by the attention weights $ \alpha_i $. This incentivizes the last hidden layer to prioritize the temporal instants according to their importance. Furthermore, it is possible to analyze the attention weights in order to identify the important temporal instants in the prediction process. This particularity allows the attention-based model to be more interpretable than a standard model.


\subsection{Presentation of the RETAIN Architecture}

Although the standard attention-based architecture allows some interpretability of the predictions, it still limited. Indeed, it is not possible to evaluate the importance of the input variables within a precise instant. This limitation comes from the computation of the hidden representation which is computed by a RNN. As a RNN is a non-linear model (e.g., with LSTM cells), it is non-interpretable. To overcome this limitation, Choi \textit {et al.} proposed the RETAIN architecture \cite {choi2016retain}. It separates the computation of the attention weights from the computation of the hidden representations. While the computation of attention weights is done with a recurrent neural network, the hidden representations are computed with a dense linear layer. In addition, a second recurrent neural network has been added to the RETAIN architecture to compute a second level of attention. This new attention is paid to the variable, thus allowing the model to focus on particular input variables within a specific instant. Once the attention weights have been determined, the computation of the predictions is done in a linear fashion. This makes it possible to measures the contribution of the input variables at each instant to the final prediction. The RETAIN model nonetheless remains a non-linear model thanks to the computation of attention weights being done in a non-linear way through the use of recurrent neural networks. The measurement of the contribution of each variable at each instant makes the RETAIN architecture much more interpretable than an architecture implementing the standard attention mechanism.

\begin{figure*}[!ht]
    \centering
    \includegraphics[width=\textwidth]{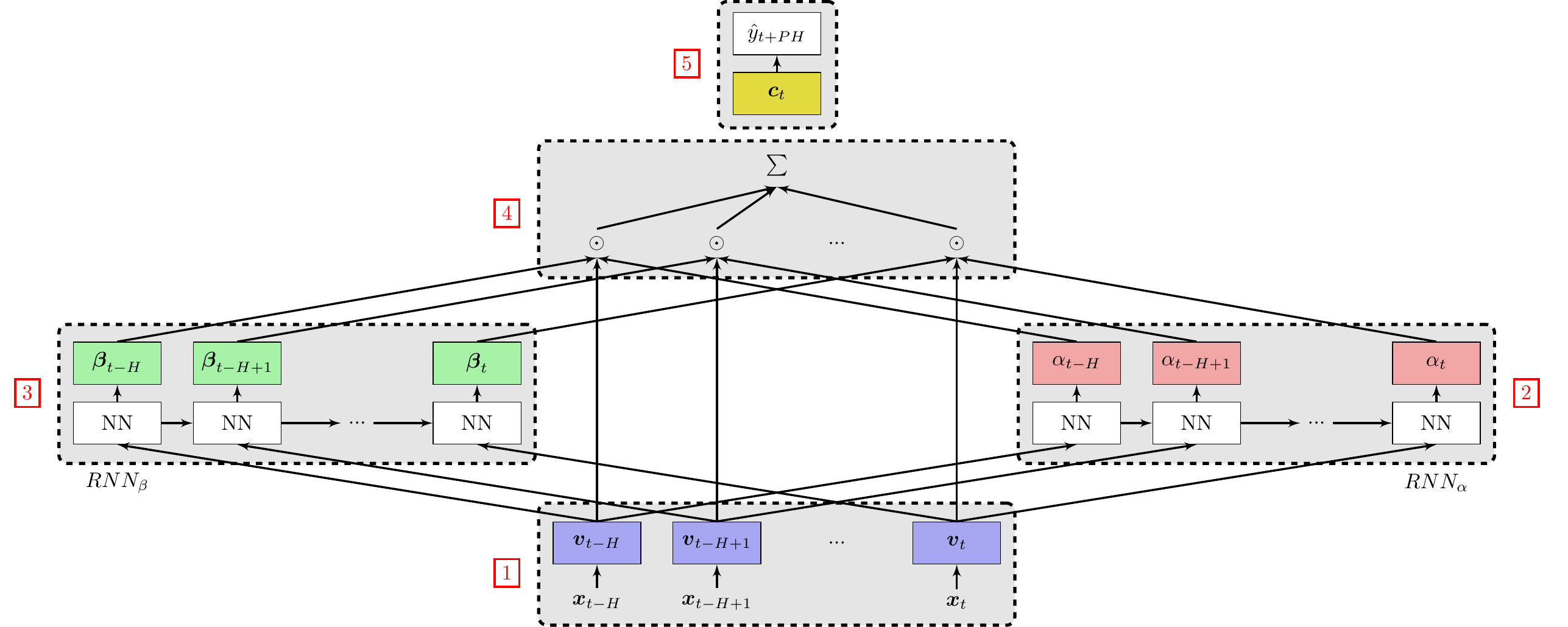}
    \caption{Graphical representation of the RETAIN model. \textbf {Step 1}: The input signals are transformed into hidden representations. \textbf {Step 2}: The weights related to temporal attention are computed from the hidden representations. \textbf {Step 3}: The weights related to the attention to the variable are also computed from the hidden representations. \textbf {Step 4}: The context vector is computed from the hidden representations weighed by the attention weights. \textbf {Step 5}: The prediction is made from the context vector.}
    \label{fig:retain}
\end{figure*}

The predictions of the RETAIN model are made in 5 steps, of which Figure \ref {fig:retain} gives a graphical representation. As before, $ \boldsymbol {x} _t \in \mathcal {R} ^ r $ represents the $ r $ input variables at time $ t $. The set of input data is represented by $ \boldsymbol {x} _ {t-H}, ..., \boldsymbol {x} _ {t} $ where $ H $ represents the length of the history known by the model.


\textbf{Step 1}: First of all, for each time instant $ i, i \in [t, t-H] $, hidden representations, also refered as embeddings in the original publication, $ \boldsymbol{v} _i \in \mathcal {R} ^ m $ are computed from the input data by the linear operation described by \ref {eqn:retain_emb} . While $ m $ represents the size of the hidden representations, $ \boldsymbol {W} _ {emb} \in \mathcal {R} ^ {m \times r} $ is the matrix allowing their computation.


\begin{equation}
    \boldsymbol{v}_i = \boldsymbol{W}_{emb} \boldsymbol{x}_i
    \label{eqn:retain_emb}
    \tag{Step 1}
\end{equation}

\textbf{Step 2}: These hidden representations are given as input to a first recurrent neural network $ \text {RNN} _ {\alpha} $ of $ p $ neurons (\ref {eqn:retain_rnna_1}, with $ g_i \in \mathcal {R} ^ p $), followed by a linear layer (\ref {eqn:retain_rnna_2}, with $ \boldsymbol {w} _ \alpha \in \mathcal {R} ^ p $ and $ b_ \alpha \in \mathcal {R} $) and Softmax normalization (\ref {eqn:retain_rnna_3}) to compute the temporal attention weights $ \alpha_i \in \mathcal {R} $. They represent the weights (positive, between 0 and 1) that the model will give at each instant $ i $ within the history to the hidden representations of the input variables. The greater the weights, the more the given instant will be taken into account in the final computation of the prediction.


\begin{subequations}
 \begin{alignat}{3}
\boldsymbol{g}_{t-H}, ..., \boldsymbol{g}_t & = \text{RNN}_\alpha(\boldsymbol{v}_{t-H}, ..., \boldsymbol{v}_t) \label{eqn:retain_rnna_1}\tag{Step 2.1}\\
e_i & = \boldsymbol{w}_\alpha^T\boldsymbol{g}_i + b_\alpha\label{eqn:retain_rnna_2}\tag{Step 2.2}\\
\alpha_{t-H}, ...,  \alpha_{t} & = \text{Softmax}(e_{t-H}, ..., e_{t}) \label{eqn:retain_rnna_3}\tag{Step 2.3}
 \end{alignat}\label{eqn:retain_rnna}
\end{subequations}

\textbf {Step 3}: Simultaneously, the features extracted in \ref {eqn:retain_emb} are also given as input to a second recurrent neural network $ \text {RNN} _ {\beta} $ of $ q $ neurons (\ref {eqn:retain_rnnb_1}). Its output, $ \boldsymbol {h} _i \in \mathcal {R} ^ q $, is used to compute the variable-level attention weights $ \boldsymbol {\beta} _i \in \mathcal {R} ^ m $ (\ref{eqn:retain_rnnb_2}, with $ \boldsymbol {W} _ \beta \in \mathcal {R} ^ {m \times q} $ and $ \boldsymbol {b} _ \beta \in \mathcal {R } ^ m $). The use of the activation function $ \tanh $ enables a positive and negative weighing, between -1 and 1, of the impact of the different embeddings from a given instant. Although the attention weights to the variable $ \boldsymbol {\beta} _i $ are directly linked to the hidden representations $ \boldsymbol {v} _i $ (see Figure \ref {fig:retain}), we can infer the attention to the variables $ \boldsymbol {x} _i $ thanks to the linearity of the computation of the hidden representations $ \boldsymbol {v} _i $.


\begin{subequations}
 \begin{alignat}{2}
\boldsymbol{h}_{t-H}, ..., \boldsymbol{h}_t & = \text{RNN}_\beta(\boldsymbol{v}_{t-H}, ..., \boldsymbol{v}_t) \label{eqn:retain_rnnb_1}\tag{Step 3.1}\\
\boldsymbol{\beta}_i & = \text{tanh}(\boldsymbol{W}_\beta\boldsymbol{h}_i + \boldsymbol{b}_\beta\label{eqn:retain_rnnb_2})\tag{Step 3.2}
 \end{alignat}\label{eqn:retain_rnnb}
\end{subequations}

\textbf{Step 4}: The context vector $ \boldsymbol {c} _t \in \mathcal {R} ^ m $ is computed as the sum, on the time axis, of the features $ \boldsymbol {v} _i $ weighted by their temporal attention $ \alpha_i $ and their attentions to the variable $ \boldsymbol {\beta} _i $ (see \ref {eqn:retain_ctx}).


\begin{equation}
\boldsymbol{c}_t = \sum_{i=t-H}^t \alpha_i\boldsymbol{\beta}_i \odot \boldsymbol{v}_i
    \label{eqn:retain_ctx}
    \tag{Step 4}
\end{equation}

\textbf{Step 5}: Finally, the prediction $ \hat {y} _ {t + PH} $ is computed by a linear dense layer according to \ref {eqn:retain_y}, where $ \boldsymbol {w} _ {out} \in \mathcal {R} ^ m $ and $ b_ {out} \in \mathcal {R} $. After computing the predictions, like any neural network, the model can adjust its different weights ($ \boldsymbol {W} _ {emb} $, $ \text {RNN} _ \alpha $, $ \text {RNN} _ \beta $, $ \boldsymbol {w} _ {\alpha} $, $ \boldsymbol {W} _ {\beta} $ and $ \boldsymbol {w} _ {out} $) by back-propagating the gradient error (e.g., mean-squared error).


\begin{equation}
    \hat{y}_{t+PH} = \boldsymbol{w}_{out}^T \boldsymbol{c}_t + b_{out}
    \label{eqn:retain_y}
    \tag{Step 5}
\end{equation}

\textbf {Differences with the RETAIN model of Choi \textit {et al.}}: The RETAIN model was initially proposed for classification tasks (e.g., detection of heart failure), tasks that are different from regression ones. Thus, we have adapted it, through \ref {eqn:retain_y}, for regression tasks.


Furthermore, in its version published by Choi \textit {et al.}, The recurrent neural networks $ \text {RNN} _ \alpha $ and $ \text {RNN} _ \beta $ process the temporal instants $ t-H $ to $ t $ in the opposite direction of time. According to the authors, this allows the model to mimic the analysis of doctors looking first on recent consultations. Our experiments did not seem to benefit from the compution of the attention weights in the opposite direction of time. Thus, we suppose that this is not a essential part of the RETAIN architecture and should be customized according to the task at hand \footnote {The name of RETAIN means \textit {REverse Time AttentIoN}. Not computing the attention weights in reverse time order makes the name of RETAIN not very adequate. Nonetheless, we have kept it to give credit to the authors.}.


\subsection{How to Interpret the RETAIN Architecture}

\subsubsection{Contribution of the input variables on the final prediction}

The coefficients $ \alpha_i $ and $ \boldsymbol {\beta} _i $ represent the weights of past temporal instants and the weights of the hidden representations $ \boldsymbol {v} _i $ in the computation of the final prediction. Thanks to its almost-linear structure, we can compute the contribution of each input variable to the prediction made by the RETAIN architecture. The Equation \ref {eqn:retain_pred_final} allows to express, from the Equation \ref {eqn:retain_y}, the computation of the final prediction $ \hat {y} _ {t + PH} $ from the input variables $ \boldsymbol {x} _ {i} $, the attentions $ \alpha_i $ and $ \boldsymbol {\beta} _i $, the matrix computing the embeddings $ \boldsymbol {W} _ {emb} $, and the one computing the final prediction $ \boldsymbol {w} _ {out} $ with its associated bias $ b_ {out} $.


\begin{subequations}
 \begin{alignat}{3}
     \hat{y}_{t+PH} &= \boldsymbol{w}_{out}^T \boldsymbol{c}_t + b_{out}\\
& = \boldsymbol{w}^T_{out}(\sum_{i=t-H}^t \alpha_i \boldsymbol{\beta}_i \odot \boldsymbol{v}_i) + b_{out}\\
  & = \boldsymbol{w}^T_{out}(\sum_{i=t-H}^t \alpha_i \boldsymbol{\beta}_i \odot (\boldsymbol{W}_{emb} \boldsymbol{x}_i)) + b_{out} \label{eqn:retain_pred_final_sub3}
 \end{alignat}
  \label{eqn:retain_pred_final} 
 \end{subequations}
 
 Equation \ref {eqn:rewrite_v} gives the rewriting of the embeddings $ \boldsymbol {v} _i $ as the sum over $ j $ of the input variables $ x_ {i, j}, j \in \mathcal  [1,r] $ weighted by the $j$-th column of the matrix $ \boldsymbol {W} _ {emb} $, $ \boldsymbol {W} _ {emb} [:, j] $.
 
 
 \begin{subequations}
  \begin{alignat}{2}
    \boldsymbol{v}_i &= \boldsymbol{W}_{emb}\boldsymbol{x}_i \\
    &= \sum_{j=1}^r {x}_{i,j}\boldsymbol{W}_{emb}[:,j]
  \end{alignat}
 \label{eqn:rewrite_v}
 \end{subequations}
 
 Starting from Equation \ref {eqn:retain_pred_final_sub3}, the computation of the final prediction $ \hat {y} _ {t + PH} $ can thus be rearranged according to Equation \ref {eqn:retain_pred_final2}.

 
 \begin{subequations}
 \begin{alignat}{3}
  \hat{y}_{t+PH} & = \boldsymbol{w}^T_{out}(\sum_{i=t-H}^t \alpha_i \boldsymbol{\beta}_i \odot (\sum_{j=1}^r {x}_{i,j}\boldsymbol{W}_{emb}[:,j])) + b_{out}\\
  & = \sum_{i=t-H}^t \sum_{j=1}^r \boldsymbol{w}^T_{out}( \alpha_i \boldsymbol{\beta}_i \odot ( {x}_{i,j}\boldsymbol{W}_{emb}[:,j])) + b_{out}\\
  & = \sum_{i=t-H}^t \sum_{j=1}^r {x}_{i,j}  \alpha_i \boldsymbol{w}^T_{out}(\boldsymbol{\beta}_i \odot \boldsymbol{W}_{emb}[:,j]) + b_{out}
  \end{alignat}
  \label{eqn:retain_pred_final2} 
 \end{subequations}
 
 This rearrangement shows that the final prediction $ \hat {y} _ {t + PH} $ is a linear combination of the input variables $ x_ {i, j} $. Thus, Equation \ref{eqn:retain_contrib_j} allows us to give a definition of the contribution, $ \omega (\hat {y} _ {t + PH}, x_ {i, j}) $, of $j$-th variable at time $ i $ on the prediction $ \hat {y} _ {t + PH} $.

 
 \begin{equation}
     \omega(\hat{y}_{t+PH}, x_{i,j}) = \underbrace{\alpha_i \boldsymbol{w}^T_{out}(\boldsymbol{\beta}_i \odot \boldsymbol{W}_{emb}[:,j])}_{\text{contribution coefficient}} \underbrace{x_{i,j}}_{\text{input variable}}
     \label{eqn:retain_contrib_j}
 \end{equation}
 
 \subsubsection{Absolute normalized contribution}
 
The contribution $ w (\hat {y} _ {t + PH}, x_ {i, j}) $ of the input variable $ x_ {i, j} $ on the prediction $ \hat {y} _ { t + PH} $ makes it possible to analyze the behavior and reasoning of the model in the computation of the final predictions. However, this value is not practical for doing statistical analysis of the average behavior of the model. First, a variable can have a negative or a positive contribution depending on the situation. Thus, the computation of the average contribution of such a variable may not be representative of its real impact on the predictions. Also, the contribution of an input variable depends on the amplitude of the prediction. Therefore, all the samples are not given the same importance in the computation of the mean contribution of an input variable to the prediction, giving more importance to high amplitude predictions. To address these limits, we propose the absolute normalized contribution, $ \omega_ {AN} (\hat {y} _ {t + PH}, x_ {i, j}) $, of the input variable $ x_ {i, j} $ on the prediction $ \hat {y} _ {t + PH} $. Described by Equation \ref {eqn:retain_an_contrib}, it allows to measure, between 0 and 1, the absolute amplitude of the contribution of the input variable $ x_ {i, j} $ on the prediction $ \hat {y} _ {t + PH} $.
 
\begin{equation}
\omega_{AN}(\hat{y}_{t+PH},x_{i,j}) = \frac{\left\lvert \omega(\hat{y}_{t+PH},x_{i,j}) \right\rvert}{\sum_{i=t-H}^t \sum_{j=1}^r \left\lvert \omega(\hat{y}_{t+PH},x_{i,j}) \right\rvert}
\label{eqn:retain_an_contrib}
\end{equation}
\section{Methods}

In this paper, we study the RETAIN model for the task of predicting future glucose values of diabetic people, task that can be described as a regression problem. From past glucose values and other information such as insulin infusions or carbohydrate (CHO) intakes, the model tries to predict the patient's future glucose values, usually between 15 to 120 minutes ahead of time \cite{oviedo2017review}.

In the section, we present the methodology that has been carried out to evaluate the RETAIN model in the context of glucose forecasting for diabetic people. First, we describe the two datasets that have been used in the study. Then, we provide the details of the implementation of the RETAIN model as well as the implementation of the reference models. Finally, we report the post-processing steps and evaluation metrics that we used.

\subsection{Experimental Data}

\subsubsection{IDIAB Dataset (I)} The IDIAB dataset has been collected by ourselves on 6 type-2 diabetic patients (5F/1M, age 56.5 $\pm$ 9.14 years old, BMI 33.52 $\pm$ 4.17 $kg/m^2$). The data collection has been approved by the French ethical commitee (ID RCB 2018-A00312-53). The patients had been monitored for 31.17 $\pm$ 1.86 days in free-living conditions. While glucose values (in $mg/dL$) have been collected using FreeStyle Libre continuous glucose monitoring devices (Abbott Diabetes Care), CHO intakes (in $g$) and insulin infusion values (in unit) have been obtained through the mySugr coaching application for diabetes.

\subsubsection{OhioT1DM Dataset (O)} The OhioT1DM dataset has been released by Marling \textit{et al.} for the  Blood Glucose Level Prediction Challenge \cite{marling2018ohiot1dm}. It is made of data coming from 6 type-1 diabetic patients (2M/4F, age between 40 and 60 years old, BMI not disclosed) that had been monitored for 8 weeks in free living conditions. To be consistent with the IDIAB dataset, in this study we only use the most important signals which are the glucose values, the insulin infusions, and the CHO intakes.

\subsection{Preprocessing}

\begin{figure}[ht]
	\includegraphics[width=\textwidth]{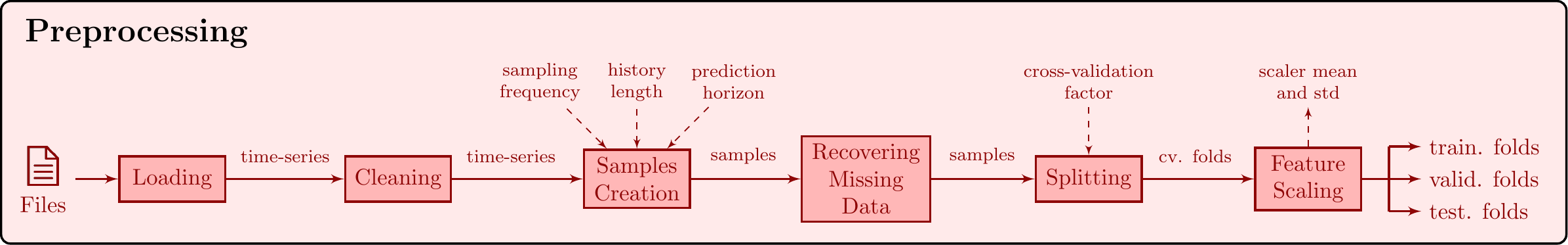}
	\centering
	\caption{Preprocessing of the data.}
	\centering
	\label{fig:preprocessing}
\end{figure}

In order to train the models efficiently, a few preprocessing steps must be carried out. Figure \ref{fig:preprocessing} gives a graphical representation of these steps. These steps are similar to the benchmark study we conducted \cite{debois2018GLYFE}.

\subsubsection{Cleaning}

The graphical analysis of glucose data from the IDIAB dataset revealed some values that appear to be erroneous. These values are characterized by peaks in blood sugar that last only for one sample (unlike the gradual increase in blood sugar following the ingestion of carbohydrates). The amount of erroneous values varies from patient to patient. Keeping them would bias the training of predictive models as well as their evaluation. Therefore, we chose to remove these values from the signals.

\subsubsection{Samples Creation}

In order to create the training samples from the glucose, CHO, and insulin signals, we need to resample them to the same sample frequency. We chose to resample the time-series to one sample every 5 minutes, which is the sampling frequency of the OhioT1DM glucose signal.

A training sample can be expressed as the set $\{\boldsymbol{X_t},y_{t+PH}\}$, where $\boldsymbol{X_t}=\boldsymbol{x}_{t-H}, ..., \boldsymbol{x}_t$ with $\boldsymbol{x}_{i}, i \in [t-H,H]$ being the input data at step $i$, and where $y_{t+PH}$ is the objective glucose value at horizon $PH$. In this study, we focus on a 30 minutes prediction horizon, as it is the most used one. As for the input values $\boldsymbol{X}_t$, they are made of the 3-hour history of glucose, CHO, and insulin values.

\subsubsection{Recovering Missing Data}

There are a lot of missing glucose values in both datasets coming from sensors or human errors. Moreover, the upsampling of the IDIAB glucose signal (from 15 minutes to 5 minutes) has also introduced a lot of missing values. Some of these values can be artificially recovered by following the following strategy for each sample:
\begin{enumerate}
    \item linearly interpolate the glucose history when the missing value is surrounded by two known glucose values;
    \item extrapolate linearly in the opposite case, usually when the missing glucose value is the most recent data;
    \item discard samples when the ground truth $y_{t+PH}$ is not known to prevent training on artificial data.
\end{enumerate}

\subsubsection{Splitting}

The OhioT1DM is originally split into training and testing sets, the testing set accounting for the last 10 days of each patient. As the IDIAB has around half as much of data, its testing sets are made of the last 5 days for each patient.

Then, every training set is split into a training and validation set following a 80\%/20\% distribution. The validation sets are used as a prior evaluation of the models when optimizing their hyperparameters. That way, the testing sets are only used for the final evaluation.

\subsubsection{Standardization}

As it is common practice in the machine-learning community, the data have been standardized (zero mean and unit variance) w.r.t. their training set.

\subsection{Glucose Predictive Models}

We present here the different glucose predictive models used in this study: the RETAIN model, the two deep reference models LSTM and FCN, as well as DT, RF and GBM, three reference models based on decision trees.

\subsubsection{Multi-Source Adversarial Transfer Learning}

To account for the high inter and intra variability of the diabetic population, glucose predictive models need to be personalized the patient \cite{oviedo2017review}. However, this reduces considerably the amount of data available, hurting the training of the models. In a previous study of ours, we showed that deep models in particular suffer from the lack of data as they are very prompt to overfit the training data \cite{debois2018GLYFE}. To alleviate this burden, we proposed in a previous study the multi-source adversarial transfer learning framework (ATL) \cite{de2020adversarial}. In the ATL setting, a first model is trained on source patients, and then finetuned to the target patient. While the target patient is the patient we want the model to be personalized to, the source patients are several patients different from the target patient. To ensure that the first model trained on the source patients generalizes well, easing the transfer to the target patient, we add an adversarial module to the initial model. Taking as input the hidden representations of the input data computed by the model, the module tries to identify the patient of origin of the given sample. When training the classifier, back-propagating the loss into the whole network, the gradient is multiplied by -1 when arriving to the computation of the hidden representation. Overall, adding the adversarial module ensures the computation of a feature representation that is useful to the task of glucose prediction but that is also patient agnostic. 

As it showed to significatively improve the accuracy of the present FCN model in our previous study, we decided to use the methodology to the LSTM and RETAIN models. In particular, we consider the intra-dataset transfer type, having the source patients being from the same dataset as the target patient. During the training on the source patients, Equation \ref{eqn:loss} expresses the loss function as the weighted combination of the mean-squared error (used for the prediction of glucose) and the multi-class cross-entropy (for the patient classification). In this equation, while $\lambda$ balances the importance of the two objectives, $\boldsymbol{y}_g, \boldsymbol{y}_p, \boldsymbol{\hat{y}}_g,\boldsymbol{\hat{y}}_p$ are respectively the glucose and patient ground truths, and the glucose and patient predictions.

\begin{equation}
\text{Loss}(\boldsymbol{y}_g, \boldsymbol{y}_p, \boldsymbol{\hat{y}}_g,\boldsymbol{\hat{y}}_p) = \text{MSE}(\boldsymbol{y}_g, \boldsymbol{\hat{y}}_g) + \lambda \cdot \text{Cross-Entropy}(\boldsymbol{y}_p, \boldsymbol{\hat{y}}_p)
    \label{eqn:loss}
\end{equation}

As for the other reference models based on decision trees, as they are not neural networks, we could not use the transfer learning methodology. As a consequence, these models are directly trained on the individual patients.

\subsubsection{RETAIN model}

The RETAIN architecture has three different elements to configure: the dimension of the extracted features $ \boldsymbol {v} _i $ and the sizes and natures of the recurrent neural networks RNN$_\alpha$ and RNN$_ \beta $. After a grid search on the validation set, we chose a feature dimension of 64 as well as recurrent networks of LSTM nature with a single layer of 128 units.

In order to implement the multi-source adversarial transfer learning  methodology, we have added to the RETAIN architecture a patient classifier module. It has been positioned after the computation of the context vector $ \boldsymbol {c} _t $ which represents the final hidden representation used for the prediction. Symmetrically with the computation of the glucose prediction being done with a dense layer, the patient classification is done with a dense layer followed by a Softmax normalization. This allows the patient classifier module to be trained to minimize multi-class cross-entropy, the error gradient of which is reversed when arriving at the computation of the context vector.

The training of the RETAIN model was done using the Adam optimizer and by mini-batch of 50 samples. The overall learning rate was $ 10 ^ {- 3} $ when training on the source patients, then $ 10 ^ {- 4} $ when finetuning the model on the target patient. To avoid overfitting the model to training data, the early stopping methodology was used with a patience of 100 epochs when training on source patients and 25 epochs when finetuning to the target patient. Finally, the coefficient $ \lambda $ was $ 10 ^ {- 2.5} $, maximizing the MSE obtained after transfer on the validation set of the target patient.

\subsubsection{Deep reference models}

The RETAIN model uses LSTM recurrent neural networks to calculate attention weights. In order to evaluate the performances linked to this particular use of the LSTM network, we can use a standard LSTM model. Like the RETAIN model, the LSTM model can use the adverse transfer learning. For this, we can link the hidden representation of the network, usually linked to a dense layer to make the glucose prediction, to a second parallel dense layer performing the patients classification. Similarly, the patient classifier is trained to minimize multi-class cross-entropy, the error gradient of which is reversed upon arriving at the LSTM network. In this study, we use the architecture and training hyperparameters of the LSTM model of the benchmark study we conducted \cite{debois2018GLYFE}. It consists of two layers of 256 LSTM units. It is trained with the Adam optimizer by mini-batch of 50 samples with a learning rate of $ 10 ^ {- 3} $ during training on source patients, and of $ 10 ^ {- 4} $ during finetuning on the target patient. A L2 regularization of $ 10 ^ {- 4} $ as well as the early stopping methodology (patience of 100 epochs while learning on the source patients, then 25 on the target patient) were used to limit the overfitting of the model. Finally, as for the RETAIN model, the gradient of the error linked to the multi-class cross-entropy of the patient classifier is weighted by $ \lambda = 10 ^ {- 2.5} $.

Also, we include the FCN model that has been used when studying the multi-source adversarial transfer learning methodology \cite{de2020adversarial}. The hidden representation is computed by 3 convolutional layers (1-dimensional convolution of size 3 $\rightarrow$ ReLU activation function $\rightarrow$ batch normalization $\rightarrow$ dropout) with 64, 128, and 64 channels respectively. From the hidden representation, the glucose prediction is computed as a dense layer of 2048 neurons and the patient probability distribution is computed with another parallel dense layer of 2048 neurons. The FCN has been trained with the Adam optimizer by mini-batch of size 100, a learning of $10^{-4}$ when training on the source patients and of $10^{-5}$ during finetuning. For regularisation, we used a dropout rate of 50\% and the early stopping methodology (patience of 250 first, and then 50). The MSE and cross-entropy losses have been weighted by $\lambda=10{-0.75}$.

\subsubsection{Reference models based on decision trees}

In order to evaluate the performances of the RETAIN model, we also chose to compare it with a simple but interpretable decision tree (DT) model. We complete this model with two other models, random forests (RF) and gradient boosting machines (GBM), both based on sets of decision trees. The RF and GBM models are generally more efficient than simple decision trees thanks to their complexity. While this performance gain comes with a drop in interpretability, these models are still more interpretable than most machine learning models by being able to measure the Gini importance of the input variables. For a single tree, the Gini importance of an input variable is computed as the drop in impurity by the node making the decision on this variable, weighted by the probability of reaching the node. For an RF or GBM model, the importance of the variables is averaged over the entire forest.

The DT model is a standard decision tree. Although simple in nature, decision trees have been used several times for the blood glucose prediction task \cite{mayo2019glycemic,li2016smartphone}. When creating the tree, in order to reduce the impact of overfitting on training data, we can constrain a branch separation to have a minimum number of training samples supporting this separation. This number has been set at 100 for the IDIAB dataset and 500 for the OhioT1DM dataset. This difference can be explained by a higher total number of training samples for the OhioT1DM dataset, thus allowing a stronger constraint on the branch separation.


 The random forest (RF) model, is an ensemble model based on decision trees. It is composed of a large number of decision trees, each tree being different from the others thanks to a randomization process used during their creation. This randomization affects both the input variables used when creating new branches, but also the selection of samples used for their creation. This randomization encourages diversity within the forest, allowing the final prediction, computed as the average of the individual decisions, to be more accurate. More efficient than traditional decision trees, randomized forests are increasingly being used for the task of blood glucose prediction \cite{midroni2018predicting,jeon2019predicting,mayo2019glycemic,georga2012predictive}. In this study, we used a forest of 100 trees. As for the DT model, we optimized by grid search the constraint of the minimum number of samples for having branch separation. This value was set to 50 and 250 samples for the IDIAB and OhioT1DM sets respectively. We note that these values are lower than for the DT model. This is intuitively explained by a lower need for regularization, which is already partly performed by the forest creation mechanism.

The GBM model is built around the gradient boosting technique. Iteratively, decision trees are created, each tree having the objective of reducing the errors of the previously created trees. This method differs from random forests where trees are created simultaneously. Like random forests, models based on gradient boosting (e.g., GBM, XGBoost) are also increasingly used in the field of blood glucose prediction : \cite{midroni2018predicting,jeon2019predicting,mayo2019glycemic}. As for the DT and RF models, we have optimized the minimum number of samples required to create new branches to 250 and 2000 for the IDIAB and OhioT1DM datasets respectively. These significantly higher values induce shallower trees, which is common for GBM models. During the iterative creation of the trees, the contribution of each tree to the final prediction is decreased by a coefficient called the learning rate. In this study, we optimized the learning rate to a value of $10^{-1}$. Also, we stopped the training after 10 iterations without performance improvement on the validation subset. This method is similar to the early stopping method used in deep learning.


\subsection{Evaluation of the predictive models}

\begin{figure}[ht]
	\includegraphics[width=\textwidth]{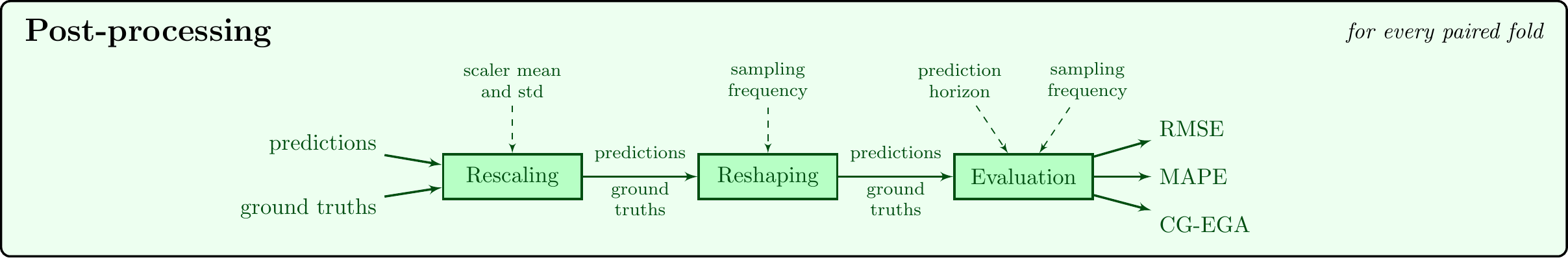}
	\centering
	\caption{Post-processing et evaluation of the predictions.}
	\centering
	\label{fig:retain_postprocessing}
\end{figure}

The evaluation of the predictive models is done following the steps described by Figure \ref{fig:retain_postprocessing}. In this study we focus on the 30-minutes prediction horizon. Before evaluating the predictions, we follow two post-processing steps. First, we rescale the predictions to their original scale (see the \textit{features scaling} preprocessing step). Then, we reconstruct the prediction time-series by reordering the predictions.

To evaluate the models we use three different metrics: the RMSE, the MAPE, and the CG-EGA. For each metric, the performances are averaged over the 5 test subsets of each patient linked to the 5-fold cross-validation, then all the patients of the same data set. Both the RMSE and MAPE metrics give a complementary measure of the accuracy of the prediction. While the RMSE is closely related to the actuel prediction scale, the MAPE is scale independent and is expressed in percentage. On the other hand, the CG-EGA measures the clinical acceptability of the prediction by analyzing the clinical accuracy as well as the coherence between successive predictions \cite{kovatchev2004evaluating}. In the end, the CG-EGA classifies a prediction either as an accurate prediction (AP), a benign error (BE), or an erroneous prediction (EP). A high AP rate and a low EP rate are necessary for a model to be clinically acceptable. The rates can be either averaged over all the test samples, or for the samples within a specific glycemic region (i.e., hypoglycemia, euglycemia and hyperglycemia).

\section{Results \& Discussion}

In this section, we first present the statistical and clinical results of the various models we presented. Then, we analyze and interpret the predictions made by the RETAIN model.

\subsection{Presentation of the experimental results}

\begin{table}[!ht]
\footnotesize
\centering
\caption{Mean (with standard deviation) of statistical accuracy (RMSE and MAPE) and general clinical acceptability (CG-EGA) for a prediction horizon of 30 minutes and for the IDIAB and OhioT1DM datasets.}
\label{table:retain_general_results}
\begin{center}

\begin{tabular}{c||c|c|ccc}
\toprule

\multirow{2}{*}{\textbf{Model}} &\multirow{2}{*}{\textbf{RMSE}} &\multirow{2}{*}{\textbf{MAPE}} &   \multicolumn{3}{c}{\textbf{CG-EGA (general)}} \\

&&& AP & BE & EP \\

\midrule
\multicolumn{6}{c}{\textbf{\textit{IDIAB dataset}}}\\
\midrule

\textbf{DT} & 24.45 \scriptsize{(6.69)} & 11.44 \scriptsize{(1.58)} & 88.18 \scriptsize{(4.87)} & 8.38 \scriptsize{(2.77)} & 3.44 \scriptsize{(2.38)}\\

\textbf{RF} & 22.35 \scriptsize{(6.33)} & 10.33 \scriptsize{(1.50)} & 92.15 \scriptsize{(4.51)} & \underline{\textbf{4.76 \scriptsize{(2.70)}}} & 3.09 \scriptsize{(2.12)}\\

\textbf{GBM} & 21.97 \scriptsize{(6.13)} & 10.13 \scriptsize{(1.60)} & 91.80 \scriptsize{(4.29)} & 5.05 \scriptsize{(2.53)} & 3.15 \scriptsize{(2.13)}\\

\midrule

\textbf{LSTM} & 19.27 \scriptsize{(5.93)} & 8.66 \scriptsize{(1.00)} & 92.12 \scriptsize{(2.90)} & 5.57 \scriptsize{(1.56)} & \underline{\textbf{2.31 \scriptsize{(1.69)}}}\\

\textbf{FCN} & \underline{\textbf{18.51 \scriptsize{(5.48)}}} & \underline{\textbf{8.44 \scriptsize{(1.07)}}} & 92.23 \scriptsize{(3.57)} & 5.27 \scriptsize{(2.09)} & 2.50 \scriptsize{(2.00)}\\

\midrule
\textbf{RETAIN} & 19.49 \scriptsize{(5.69)} & 8.71 \scriptsize{(0.75)} & \underline{\textbf{92.41 \scriptsize{(2.94)}}} & 5.15 \scriptsize{(1.60)} & 2.43 \scriptsize{(1.58)}\\

\midrule
\multicolumn{6}{c}{\textbf{\textit{OhioT1DM dataset}}}\\
\midrule

\textbf{DT} & 23.87 \scriptsize{(2.28)} & 11.22 \scriptsize{(2.54)} & 79.07 \scriptsize{(3.92)} & 16.81 \scriptsize{(2.40)} & \underline{\textbf{4.12 \scriptsize{(2.13)}}}\\

\textbf{RF} & 22.03 \scriptsize{(2.41)} & 10.14 \scriptsize{(2.38)} & \underline{\textbf{83.67 \scriptsize{(4.01)}}} & \underline{\textbf{11.89 \scriptsize{(2.22)}}} & 4.44 \scriptsize{(2.28)}\\

\textbf{GBM} & 21.43 \scriptsize{(2.35)} & 9.78 \scriptsize{(2.48)} & 83.09 \scriptsize{(3.85)} & 12.07 \scriptsize{(1.82)} & 4.84 \scriptsize{(2.38)}\\

\midrule

\textbf{LSTM} & 19.68 \scriptsize{(2.45)} & 8.81 \scriptsize{(2.23)} & 79.37 \scriptsize{(4.51)} & 15.61 \scriptsize{(3.33)} & 5.02 \scriptsize{(1.96)}\\

\textbf{FCN} & \underline{\textbf{19.27 \scriptsize{(1.78)}}} & \underline{\textbf{8.68 \scriptsize{(1.97)}}} & 78.73 \scriptsize{(4.59)} & 15.96 \scriptsize{(3.04)} & 5.31 \scriptsize{(2.17)}\\

\midrule
\textbf{RETAIN} & 20.29 \scriptsize{(2.40)} & 9.16 \scriptsize{(2.24)} & 80.98 \scriptsize{(4.84)} & 14.28 \scriptsize{(3.22)} & 4.74 \scriptsize{(2.17)}\\

\bottomrule
\end{tabular}

\end{center}

\begin{flushright}
AP: {Accurate Prediction}; BE:{ Benign Error}; EP: {Erroneous Prediction}
\end{flushright}

\end{table}{}

Table \ref {table:retain_general_results} presents the mean precision (RMSE and MAPE) as well as the general clinical acceptability (general CG-EGA) of the models DT, RF, GBM, LSTM, FCN and RETAIN for the IDIAB and OhioT1DM datasets. Table \ref {table:retain_cgega_results} details, for each glycemic region, the clinical acceptability of the models (CG-EGA by region).

\setlength{\tabcolsep}{4.5pt}

\begin{landscape}
\topskip0pt
\vspace*{\fill}
\begin{table}[htbp!]
\centering
\begin{center}
\caption{Mean (with standard deviation) of per-region clinical acceptability (CG-EGA) for a prediction horizon of 30 minutes and for the IDIAB and OhioT1DM datasets.}
\footnotesize
\label{table:retain_cgega_results}
\begin{tabular}{@{}c||*{3}{c}|*{3}{c}|*{3}{c}@{}}
\toprule

\multirow{3}{*}{\textbf{Model}} &    \multicolumn{9}{c}{\textbf{CG-EGA (by region)}} \\

& \multicolumn{3}{c|}{\textit{Hypoglycemia}} & \multicolumn{3}{c|}{\textit{Euglycemia}} & \multicolumn{3}{c}{\textit{Hyperglycemia}} \\ 

& AP & BE & EP & AP & BE & EP & AP & BE & EP \\

\midrule
\multicolumn{10}{c}{\textbf{\textit{IDIAB dataset}}}\\
\midrule
\textbf{DT} & 36.49 \scriptsize{(27.43)} & 0.57 \scriptsize{(1.14)} & 62.94 \scriptsize{(27.77)} & 92.47 \scriptsize{(1.98)} & 6.65 \scriptsize{(1.36)} & 0.88 \scriptsize{(0.63)} & 85.07 \scriptsize{(8.13)} & 11.62 \scriptsize{(4.79)} & 3.30 \scriptsize{(3.54)}\\

\textbf{RF} & 33.10 \scriptsize{(29.94)} & \underline{\textbf{0.00 \scriptsize{(0.00)}}} & 66.90 \scriptsize{(29.94)} & \underline{\textbf{96.38 \scriptsize{(1.54)}}} & \underline{\textbf{3.10 \scriptsize{(1.33)}}} & 0.52 \scriptsize{(0.37)} & 89.45 \scriptsize{(7.38)} & 7.53 \scriptsize{(4.69)} & 3.02 \scriptsize{(2.78)}\\

\textbf{GBM} & 31.81 \scriptsize{(29.18)} & 1.14 \scriptsize{(2.29)} & 67.05 \scriptsize{(28.58)} & 95.86 \scriptsize{(1.60)} & 3.58 \scriptsize{(1.39)} & 0.56 \scriptsize{(0.28)} & 88.96 \scriptsize{(6.81)} & 7.84 \scriptsize{(4.08)} & 3.21 \scriptsize{(2.85)}\\

\midrule

\textbf{LSTM} & 52.02 \scriptsize{(30.67)} & \underline{\textbf{0.00 \scriptsize{(0.00)}}} & 47.98 \scriptsize{(30.67)} & 95.17 \scriptsize{(1.41)} & 4.45 \scriptsize{(1.46)} & \underline{\textbf{0.37 \scriptsize{(0.35)}}} & \underline{\textbf{89.63 \scriptsize{(5.60)}}} & 7.65 \scriptsize{(3.25)} & 2.72 \scriptsize{(2.49)}\\

\textbf{FCN} & 51.84 \scriptsize{(30.57)} & \underline{\textbf{0.00 \scriptsize{(0.00)}}} & 48.16 \scriptsize{(30.57)} & 95.87 \scriptsize{(1.27)} & 3.62 \scriptsize{(1.15)} & 0.51 \scriptsize{(0.57)} & 88.82 \scriptsize{(5.99)} & 8.38 \scriptsize{(3.91)} & \underline{\textbf{2.81 \scriptsize{(2.64)}}}\\

\midrule
\textbf{RETAIN} & \underline{\textbf{57.09 \scriptsize{(33.07)}}} & \underline{\textbf{0.00 \scriptsize{(0.00)}}} & \underline{\textbf{42.91 \scriptsize{(33.07)}}} & 95.63 \scriptsize{(1.42)} & 3.94 \scriptsize{(1.47)} & 0.43 \scriptsize{(0.52)} & 89.09 \scriptsize{(5.39)} & \underline{\textbf{7.40 \scriptsize{(3.03)}}} & 3.51 \scriptsize{(2.65)}\\

\midrule
\multicolumn{10}{c}{\textbf{\textit{OhioT1DM dataset}}}\\
\midrule

\textbf{DT} & 23.67 \scriptsize{(13.57)} & 3.61 \scriptsize{(2.05)} & 72.72 \scriptsize{(14.98)} & 80.96 \scriptsize{(4.11)} & 16.60 \scriptsize{(3.06)} & \underline{\textbf{2.44 \scriptsize{(1.15)}}} & 79.51 \scriptsize{(2.71)} & 17.65 \scriptsize{(2.06)} & \underline{\textbf{2.84 \scriptsize{(1.27)}}}\\

\textbf{RF} & 25.51 \scriptsize{(17.82)} & \underline{\textbf{1.42 \scriptsize{(1.57)}}} & 73.07 \scriptsize{(18.34)} & 86.61 \scriptsize{(3.72)} & 10.82 \scriptsize{(2.79)} & 2.57 \scriptsize{(1.11)} & \underline{\textbf{82.53 \scriptsize{(3.26)}}} & \underline{\textbf{13.92 \scriptsize{(2.37)}}} & 3.55 \scriptsize{(1.71)}\\

\textbf{GBM} & 26.60 \scriptsize{(19.79)} & 1.74 \scriptsize{(1.87)} & 71.65 \scriptsize{(20.89)} & \underline{\textbf{86.73 \scriptsize{(3.43)}}} & \underline{\textbf{10.33 \scriptsize{(2.49)}}} & 2.93 \scriptsize{(1.15)} & 80.69 \scriptsize{(4.16)} & 15.00 \scriptsize{(2.66)} & 4.31 \scriptsize{(2.01)}\\

\midrule

\textbf{LSTM} & \underline{\textbf{46.31 \scriptsize{(24.61)}}} & 2.43 \scriptsize{(3.62)} & \underline{\textbf{51.25 \scriptsize{(25.13)}}} & 83.02 \scriptsize{(5.57)} & 13.48 \scriptsize{(4.49)} & 3.50 \scriptsize{(1.28)} & 75.96 \scriptsize{(4.03)} & 18.74 \scriptsize{(3.38)} & 5.30 \scriptsize{(1.89)}\\

\textbf{FCN} & 44.98 \scriptsize{(30.20)} & 2.83 \scriptsize{(2.75)} & 52.19 \scriptsize{(30.09)} & 82.29 \scriptsize{(5.59)} & 13.99 \scriptsize{(4.25)} & 3.71 \scriptsize{(1.48)} & 75.35 \scriptsize{(3.89)} & 18.86 \scriptsize{(3.11)} & 5.78 \scriptsize{(1.90)}\\

\midrule

\textbf{RETAIN} & 44.08 \scriptsize{(23.77)} & 2.89 \scriptsize{(2.91)} & 53.03 \scriptsize{(24.80)} & 84.11 \scriptsize{(6.14)} & 12.57 \scriptsize{(4.64)} & 3.33 \scriptsize{(1.66)} & 78.81 \scriptsize{(3.10)} & 16.58 \scriptsize{(2.46)} & 4.61 \scriptsize{(1.78)}\\

\bottomrule

\end{tabular}

\end{center}

\begin{flushright}
AP: {Accurate Prediction}; BE:{ Benign Error}; EP: {Erroneous Prediction}
\end{flushright}
\end{table}
\vspace*{\fill}
\end{landscape}

First of all, within the reference models based on decision trees (DT, RF, and GBM), we can  observe the low precision and clinical acceptability of the DT model in comparison with the RF and GBM models. This is not surprising and is explained by the simplicity of a simple decision tree. Between the RF and GBM models, the GBM model has a better statistical accuracy (RMSE and MAPE) but also a poorer clinical acceptability (AP, BE and EP scores for all regions of the CG-EGA). Overall, the results for the DT, RF and GBM models are similar for the two datasets. The very good scores in benign BE error percentages of the RF model show that it is capable of producing successive predictions that are consistent with each other. Indeed, a prediction is characterized as BE when it is clinically sufficiently accurate, but the rate of change from the previous prediction is not. A model with a high BE rate is generally a model showing high amplitude oscillations in its successive predictions.

As for the deep reference models LSTM and FCN, they show performances (precision and clinical acceptability) highly superior to the models based on decision trees. Only the clinical acceptability in the region of euglycemia and hyperglycemia for the OhioT1DM game is less good than the ones of the RF and GBM models (lower AP and higher EP rates). We note that the performance of the LSTM model in this study improved upo,n the LSTM results of our benchmark study \cite{debois2018GLYFE}. This improvement in accuracy comes from the use of the multi-source adversarial transfer learning methodology.

The RETAIN model shows a compromise between accuracy and interpretability. Indeed, the it is clearly more accurate than models based on decision trees while remaining interpretable. However, its accuracy remains slightly lower than the LSTM or FCN models. We can attribute this difference to the relative simplicity of the computation of the prediction by RETAIN. In the RETAIN architecture, the non-linearity of the computation resides only in the computation of the attention weights. This forces the extracted features to keep a certain simplicity. From the point of view of the clinical acceptability, the RETAIN model is slightly better, if not equivalent, than the FCN and LSTM models.

\subsection{Interpretability of the RETAIN model}

\begin{figure}[!ht]
	\includegraphics[width=\textwidth]{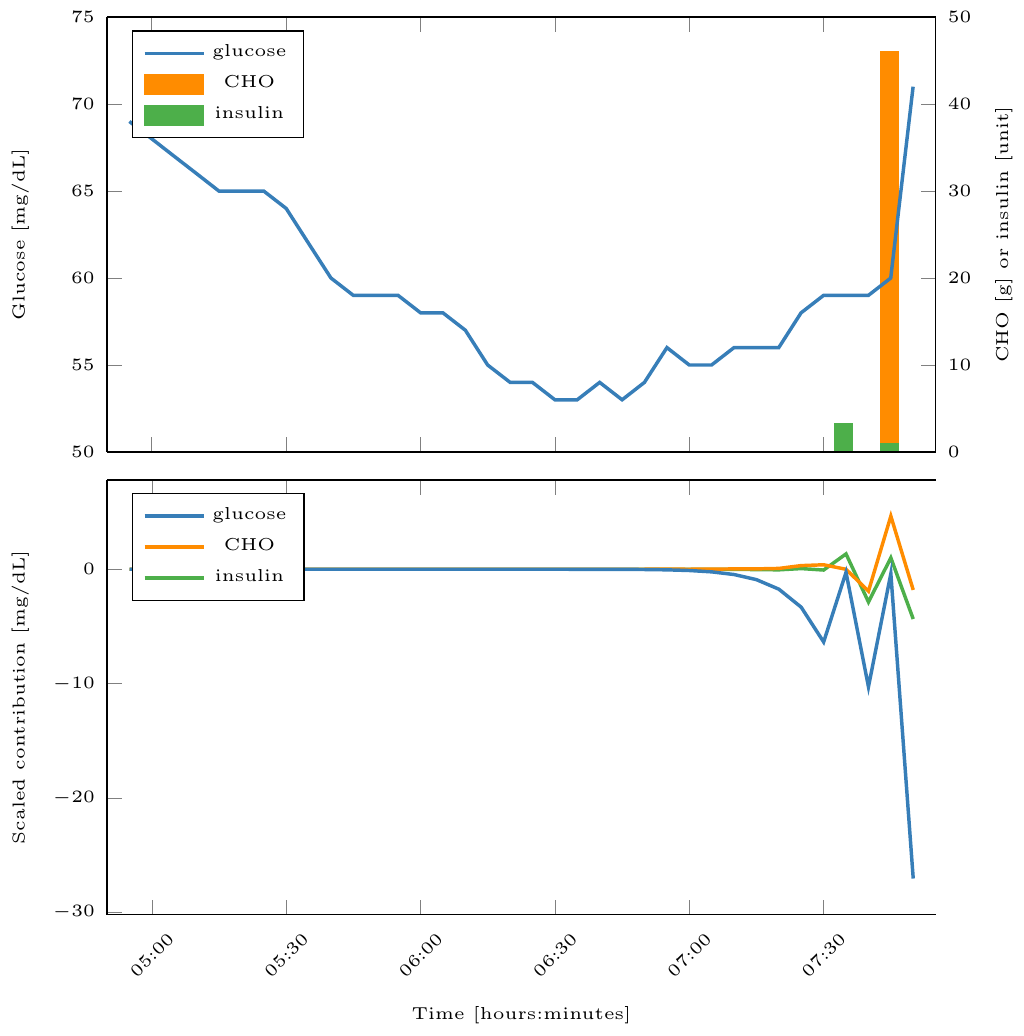}
	\centering
	\caption{Input variables of a test sample of patient 575 from the OhioT1DM set (top) and contribution of the variables to the prediction made by the RETAIN model (bottom).}
	\centering
	\label{fig:retain_example}
\end{figure}

The greatest strength of the RETAIN model lies in its interpretability. Indeed, by measuring the contribution, it is possible to quantify the impact of each variable to the prediction, thus lifting the veil on the reasoning of the model. Figure \ref {fig:retain_example} gives an example of this ability. In this example, we see that the variables with the greatest impact on the prediction are recent glucose readings. They have a significant contribution up to 1 hour in the past (1 hour history). When it comes to the CHO intakes and insulin injections, we can see contribution peaks when they appear. At these same times, the contribution of the glucose signal is close to zero. This is made possible by the attention to the variable $ \boldsymbol {\beta} _i $ computed by RNN$_\beta $. Indeed, the sole presence of the temporal attention $\alpha_i$ computed by RNN$_\alpha $ would not have made it possible to attribute a strong contribution to insulin or carbohydrate signals and simultaneously a weak contribution to the glucose signal. Finally, on this example, we can note that the contribution of variables older than one hour is close to zero.

\begin{figure}[!ht]
\centering
\begin{subfigure}{.5\textwidth}
  \centering
  \includegraphics[width=\linewidth]{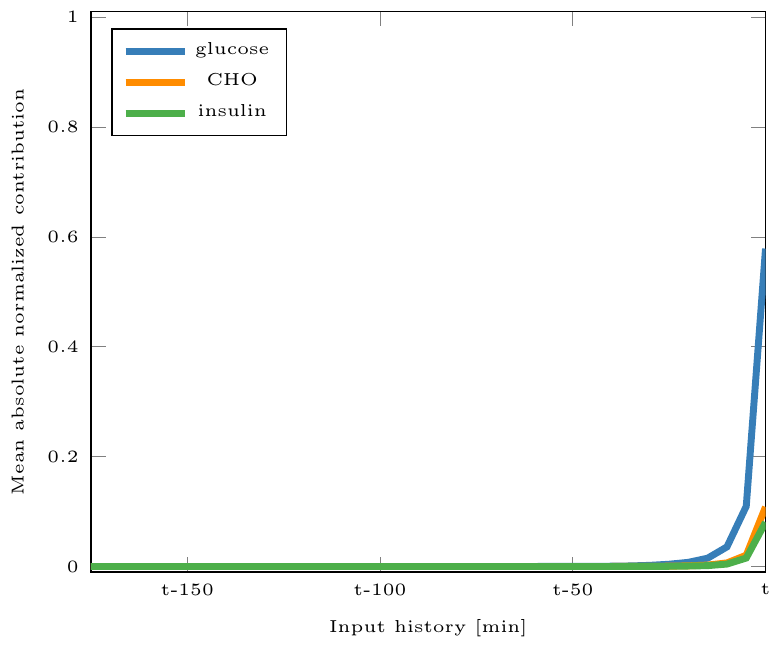}
  \caption{IDIAB dataset}
  \label{fig:distrib_idiab}
\end{subfigure}%
\begin{subfigure}{.5\textwidth}
  \centering
  \includegraphics[width=\linewidth]{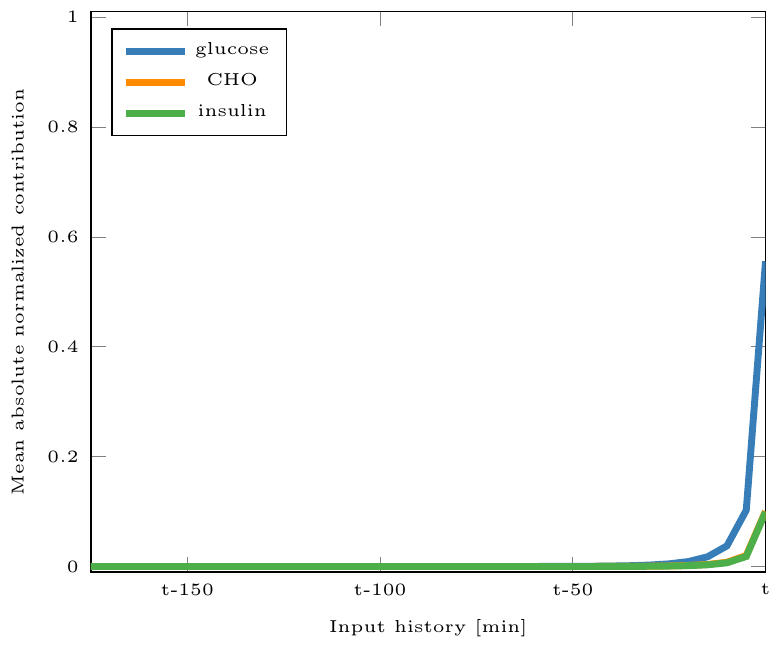}
  \caption{OhioT1DM dataset}
  \label{fig:distrib_ohio}
\end{subfigure}
\caption{Mean absolute normalized contribution of the input variables for the patients of the IDIAB (left) and OhioT1DM (right) datasets.}
\label{fig:mean_abs_contrib}
\end{figure}

\begin{figure}[!ht]
\centering
\begin{subfigure}{.5\textwidth}
  \centering
  \includegraphics[width=\linewidth]{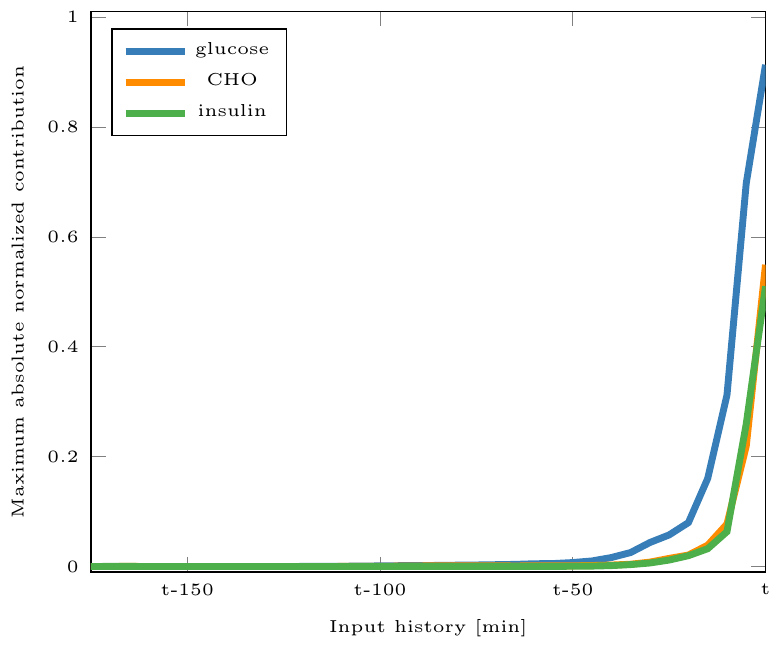}
  \caption{IDIAB dataset}
  \label{fig:distrib_idiab}
\end{subfigure}%
\begin{subfigure}{.5\textwidth}
  \centering
  \includegraphics[width=\linewidth]{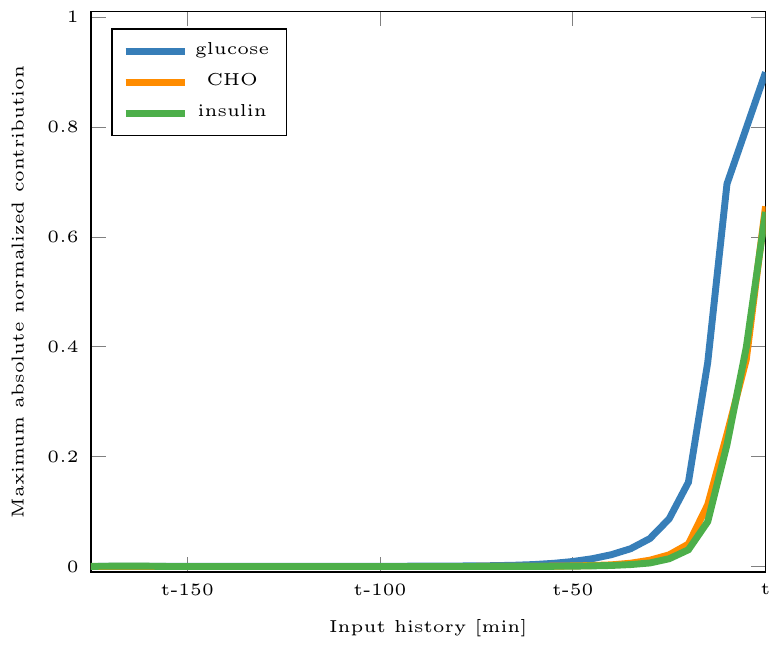}
  \caption{OhioT1DM dataset}
  \label{fig:distrib_ohio}
\end{subfigure}
\caption{Maximum absolute normalized contribution of the input variables for the patients of the IDIAB (left) and OhioT1DM (right) datasets.}
\label{fig:max_abs_contrib}
\end{figure}

We can use the mean and maximum normalized absolute contribution of each variable at any time to assess their overall usefulness for predicting future glucose values. While the mean contribution is used to analyze average behavior, the maximum contribution is used to assess whether a variable was useful at least once for all test samples. Indeed, if a variable has been useful at least once, then its maximum normalized absolute contribution will be high (equal to its usefulness). Conversely, if a variable is not used by the model to compute the predictions, then its contribution will be close to zero. Figures \ref {fig:mean_abs_contrib} and \ref {fig:max_abs_contrib} respectively represent the mean and maximum normalized absolute contribution of each variable for the IDIAB and OhioT1DM datasets. First, we can see that the interest of each signal decreases with how old it is. The older a variable, the less it contributes to predictions. This decrease is faster for the CHO and or insulin signals than for the glucose signal. While the CHO and insulin signals are no longer of interest after about 40 minutes, the glucose signal continues to impact predictions for up to 60 minutes. Beyond 60 minutes, no variable shows to be of interest for the forecasting of glucose. However, the other models studied in previous studies of ours have shown to benefit from a history longer an hour \cite{debois2018GLYFE,de2020adversarial}. This suggests that the RETAIN model is not able to efficiently use such a long history. This limitation would explain the slightly poorer performances of the model compared to the LSTM and FCN models. Moreover, the comparison of the mean normalized absolute contribution of the two IDIAB and OhioT1DM datasets through Figure \ref {fig:mean_abs_contrib} shows that the variables of the two datasets, despite their intrinsic differences (type of diabetes, material and experimental protocol), behave similarly within the RETAIN model.

\begin{figure}[!ht]
\centering
\begin{subfigure}{\textwidth}
  \centering
  \includegraphics[width=\linewidth]{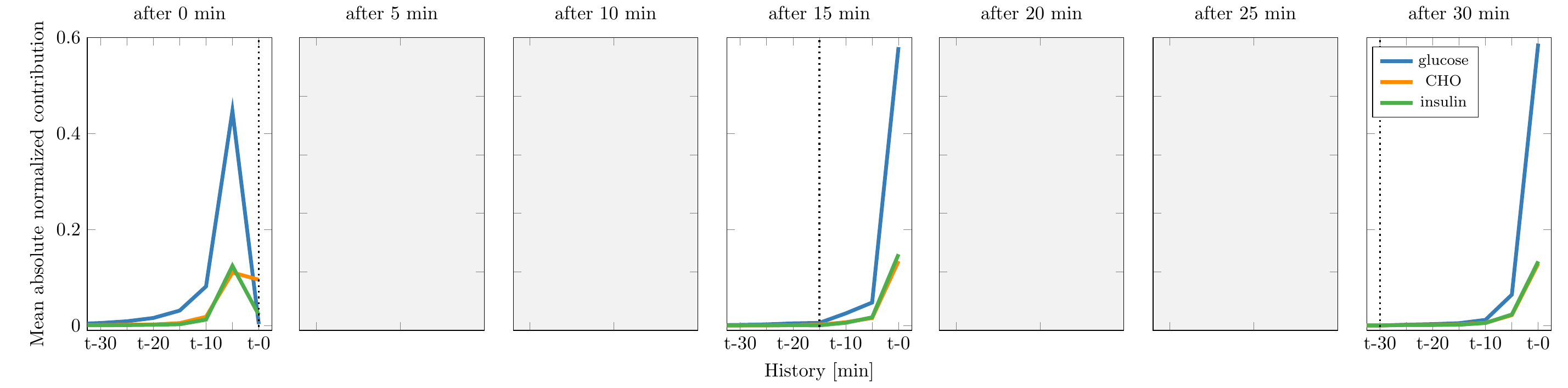}
  \caption{IDIAB - CHO ingestion}
  \label{fig:distrib_idiab}
\end{subfigure}%

\begin{subfigure}{\textwidth}
  \centering
  \includegraphics[width=\linewidth]{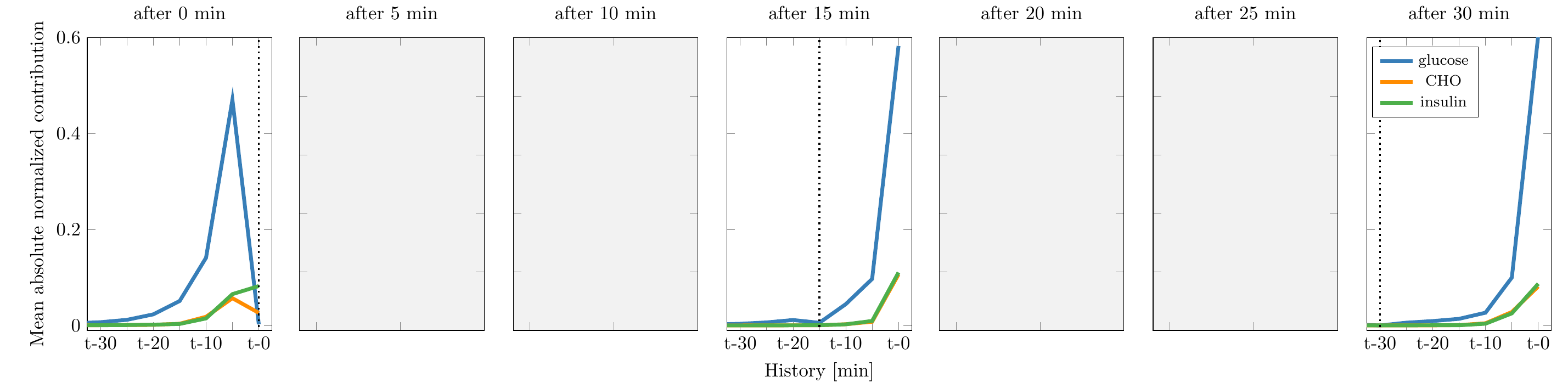}
  \caption{IDIAB - Insulin infusion}
  \label{fig:distrib_ohio}
\end{subfigure}

\begin{subfigure}{\textwidth}
  \centering
  \includegraphics[width=\linewidth]{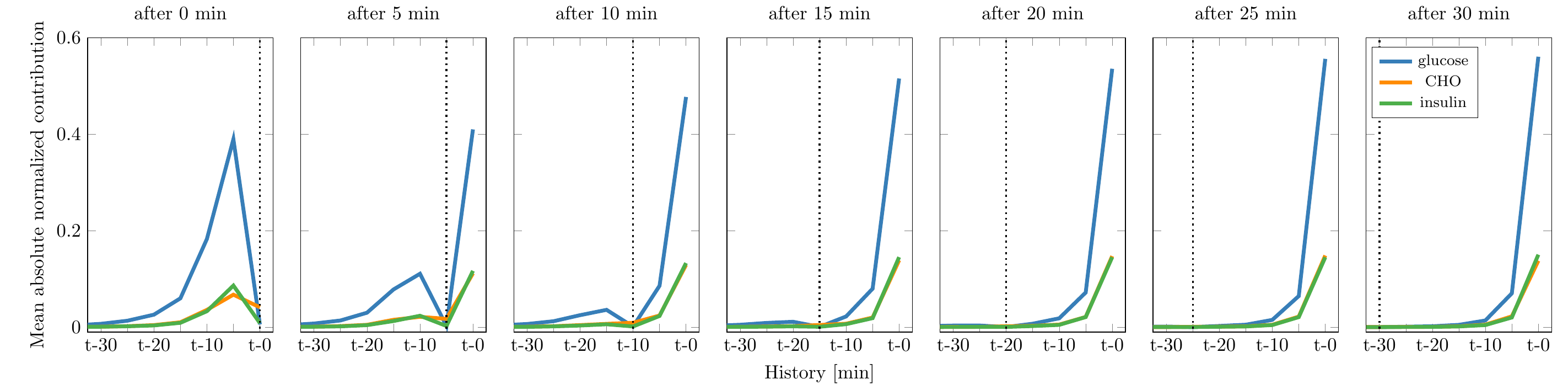}
  \caption{OhioT1DM - CHO ingestion}
  \label{fig:distrib_idiab}
\end{subfigure}%

\begin{subfigure}{\textwidth}
  \centering
  \includegraphics[width=\linewidth]{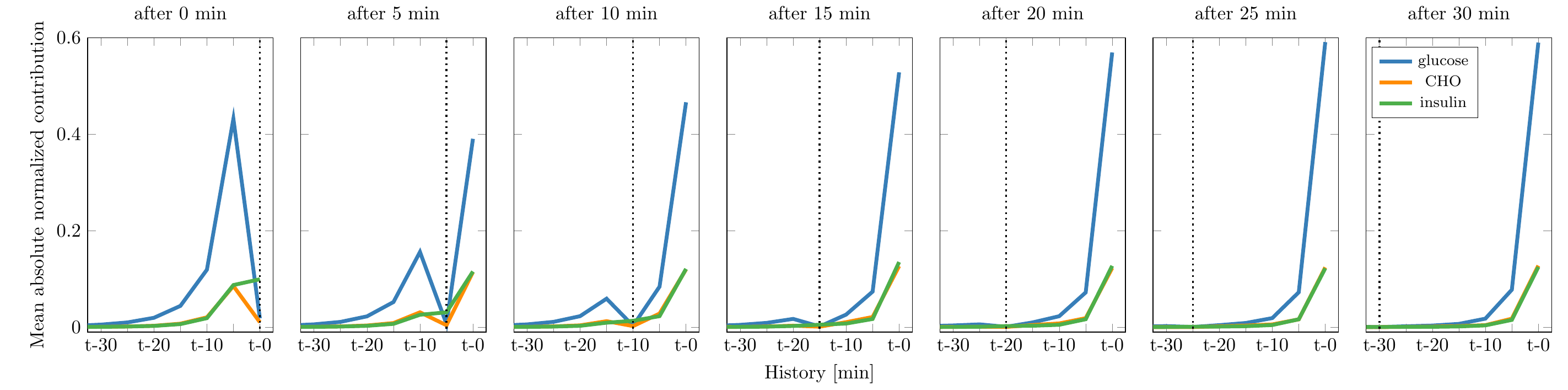}
  \caption{OhioT1DM - Insulin infusion}
  \label{fig:distrib_ohio}
\end{subfigure}
\caption{Evolution of the normalized absolute contribution of variables after an event (carbohydrate ingestion or insulin injection) averaged for the IDIAB and OhioT1DM datasets. The contribution of the IDIAB dataset are only available every 15 minutes due to the sampling frequency of the FreeStyle Libre glucose monitoring device.}
\label{fig:evolution_after_stimuli}
\end{figure}

To go further in the analysis of the contribution of each variable, we can filter the samples of interest. For example, through Figure \ref {fig:evolution_after_stimuli}, we are interested in the evolution of the contribution of the variables after the arrival of an event such as an ingestion of carbohydrates or an injection of insulin. The behavior of the model is similar both for the two types of events (CHO or insulin) and for the two datasets (IDIAB and OhioT1DM). When the event arrives, all the variables, except the one corresponding to the event, have a contribution of almost zero. It is rather the moment before the event which has a strong contribution to the prediction of glucose. Over time, although decreasing, the contribution of the moment preceding the event remains strong. This suggests that, when an event related to insulin or carbohydrate intake occurs, events that considerably modify the regulation of the patient's glycemia, the model takes into account the patient's state before the arrival of the event. After about thirty minutes, the contribution of the variables linked to the event becomes zero, indicating that the event information is no longer used by the model to make its predictions.

\section{Conclusion}

In this study, we adapted the RETAIN architecture proposed by Choi \textit {et al.} for regression tasks and analyzed its use for the prediction of future glucose values in people with diabetes. Based on neural networks, it implements a double attention mechanism allowing it to be interpretable. This ability makes it particularly interesting for biomedical tasks, and in particular for the forecasting of glucose values.

We evaluated the statistical (RMSE and MAPE) and clinical (CG-EGA) performances of the RETAIN model by comparing it to decision tree-based models and deep models. The results showed us that the models based on decision trees are largely outclassed by deep models, and in particular by the RETAIN model. In comparison with the LSTM and FCN models, the RETAIN model shows to have a slightly lower accuracy but a better, or at least equal, clinical acceptability. However, the real strength of the RETAIN model lies in its interpretability. Thanks to this ability, we carried out an analysis of the importance of the glucose, CHO and insulin signals to the forecasting of future glucose values. This analysis showed us that values older than one hour (history greater than one hour) are not used by the RETAIN model. We suppose that this limitation comes from the quasi linearity of the computation of the predictions made by the RETAIN model. We then analyzed the contribution of the input variables in the presence of an insulin infusion or CHO intake event. Following such events, the RETAIN model adopts a different behavior by strongly taking into account the moment preceding the event. After 30 minutes after the occurrence of the event, the RETAIN model returns to its standard behavior.

Overall, the RETAIN model shows to be promissing for biomedical use, and in particular for predicting future glucose values in people with diabetes. Its interpretable feature is particularly interesting both for the patient but also for the practitioners and scientists behind the creation of the model. First of all, such a model can be useful for the therapeutic education of the patient, explaining to him/her the impact of the variables on the regulation of his/her glycemia. In addition, the patient can also understand the reasoning behind the decisions of the model, and adapt his/her behavior accordingly. Finally, as we have seen in this study, the analysis of the importance of the input variables can be essential in the design of new, more efficient architectures. These new architectures can include new data from various origins, such as physical activity or sleep data. These new architectures can also be made more complex, in particular through more sophisticated hidden representations while remaining interpretable. In the original RETAIN publication, the authors discuss the use of multi-layer perceptrons for the computation of better hidden representations \cite{choi2016retain,erhan2009visualizing,le2013building}. However, this complexification of the architecture must still allow the computation of the contributions of all the variables to the final prediction in order to not lose in interpretability.

\section*{Acknowledgments}

We would like to thank the diabetes health network Revesdiab for their help in building the IDIAB dataset used in this study.

\bibliographystyle{ws-ijprai}
\bibliography{bibtex.bib}

\end{document}